\documentclass[journal,twoside,web]{ieeecolor}
\usepackage{generic}
\usepackage{cite}
\usepackage{amsmath,amssymb,amsfonts}
\usepackage{graphicx}
\usepackage{textcomp}
\usepackage{url}
\usepackage{array}
\usepackage{mathrsfs}
\usepackage{algorithm}  
\usepackage{algorithmicx}  
\usepackage{algpseudocode}  
\usepackage{bm}
\usepackage{threeparttable}
\usepackage{multirow}
\usepackage{float}

\def\BibTeX{{\rm B\kern-.05em{\sc i\kern-.025em b}\kern-.08em
    T\kern-.1667em\lower.7ex\hbox{E}\kern-.125emX}}
\begin{document}
\title{Deep Dynamic Epidemiological  Modelling for COVID-19 Forecasting in Multi-level Districts}

\author{Ruhan Liu, Jiajia Li, Yang Wen,
	Huating Li,
	Ping Zhang,\IEEEmembership{Senior Member,IEEE},
	Bin Sheng,\IEEEmembership{Member,IEEE},
	and David Dagan Feng,\IEEEmembership{Life Fellow,IEEE}
	\thanks{The source code and used datasets are already published in Github: https://github.com/Liuruhan/DENN.}	
	\thanks{R. Liu, J. Li and B. Sheng are with the Department of Computer Science and Engineering, Shanghai Jiao Tong University, Shanghai 200240, China (Email: liurh996@sjtu.edu.cn; lijiajia@sjtu.edu.cn; shengbin@sjtu.edu.cn).}
	\thanks{Yang Wen is with Shenzhen University, College of Electronics and Information Engineering, Shenzhen 518060, China (E-mail: wen$\_$yang@szu.edu.cn).}
	\thanks{H. Li is with the Shanghai Jiao Tong University Affiliated Sixth People's Hospital, Shanghai 200233, China (Email: huarting99@sjtu.edu.cn).}
	\thanks{P. Zhang is with the Department of Computer Science and Engineering, The Ohio State University, Columbus, Ohio 43210, USA; and also with the Department of Biomedical Informatics, The Ohio State University, Columbus, Ohio 43210, USA (Email: zhang.10631@osu.edu).}
	\thanks{D. D. Feng is with the School of Computer Science, The University of Sydney, Sydney, NSW 2006, Australia (Email: dagan.feng@sydney.edu.au).}}
\maketitle

\begin{abstract}
Objective: COVID-19 has spread worldwide and made a huge influence across the world. Modeling the infectious spread situation of COVID-19 is essential to understand the current condition and to formulate intervention measurements. Epidemiological equations based on the SEIR model simulate disease development. The traditional parameter estimation method to solve SEIR equations could not precisely fit real-world data due to different situations, such as social distancing policies and intervention strategies. Additionally, learning-based models achieve outstanding fitting performance, but cannot visualize mechanisms. Methods: Thus, we propose a deep dynamic epidemiological (DDE) method that combines epidemiological equations and deep-learning advantages to obtain high accuracy and visualization. The DDE contains deep networks to fit the effect function to simulate the ever-changing situations based on the neural ODE method in solving variants' equations, ensuring the fitting performance of multi-level areas. Results: We introduce four SEIR variants to fit different situations in different countries and regions. We compare our DDE method with traditional parameter estimation methods (Nelder-Mead, BFGS, Powell, Truncated Newton Conjugate-Gradient, Neural ODE) in fitting the real-world data in the cases of countries (the USA, Columbia, South Africa) and regions (Wuhan in China, Piedmont in Italy). Our DDE method achieves the best Mean Square Error and Pearson coefficient in all five areas. Further, compared with the state-of-art learning-based approaches, the DDE outperforms all techniques, including LSTM, RNN, GRU, Random Forest, Extremely Random Trees, and Decision Tree. Conclusion: DDE presents outstanding predictive ability and visualized display of the changes in infection rates in different regions and countries.
\end{abstract}


\section{Introduction}
\label{sec:introduction}
\IEEEPARstart{C}{oronavirus} Respiratory Disease 2019 (COVID-19) from the virus "SARS-CoV-2'' has a catastrophic spread and influences at least 214 countries and territories over seven continents, resulted in more than 5 million people infected and over 3456 thousand deaths. The disease has brought unprecedented impact on people's health and safety around the world and has influenced economic and social development. Due to different conditions such as quarantine measurement, social distance, population density, medical conditions, the development of the COVID-19 in different places is also diverse. In some places such as China \cite{3Cyran} and Singapore \cite{4Lin}, the infection cases appeared to be controlled in short time because many epidemic prevention measures have been popularized through propaganda. Also, adopting relatively strict quarantine measures has helped control the spread of COVID-19. The public's compliance with the quarantine measurement also is helpful. However, in others, for example, Brazil and the United States are reaching explosive growth in early time because the public is not paying enough attention to the severity of the illness. The government has not adopted strict control measures. Because of the various situations in different regions, using current data to construct an infection model to analyze the disease's condition and estimate the future development is crucial for analyzing the spread condition and helping control the outbreak of coronavirus disease.

\begin{figure}[ht]
	\begin{center}
		\includegraphics[width=3.5in]{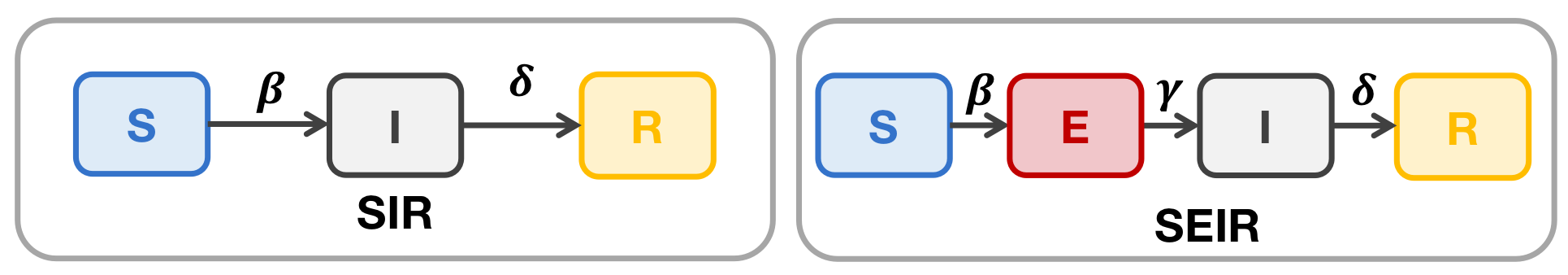}
		\caption{The architectures of SIR and SEIR models. The left figure shows the Susceptible-Infected-Removed (SIR) model. The right part shows the Susceptible-Exposed-Infected-Removed (SEIR) model.}
		\label{fig:sir}
	\end{center}
\end{figure}

Many studies propose their models to simulate the development of diseases. The Susceptible-Infected-Removed (SIR) and Susceptible-Exposed-Infected-Removed (SEIR) models are commonly used in the epidemiological analysis (Fig. \ref{fig:sir}) \cite{5Fang, 6Saito}. SIR model divides the population into three groups: the susceptible (S), the infectious (I), and the recovered (R). The S represents healthy people, and the I is those who have been infected.  Also, the R is those who have recovered from the infection. In the SEIR model modeling, the exposed group (E) who may be infected is further considered. The formula of SEIR is shown below:
\begin{equation}
	\begin{split}
		\frac{dS}{dt} = - \beta  \cdot \frac{S \cdot I}{N}\\
		\frac{dE}{dt} = \beta \cdot \frac{S \cdot I}{N} - \gamma \cdot E\\
		\frac{dI}{dt} = \gamma \cdot E - \delta \cdot I\\
		\frac{dR}{dt} = \delta \cdot I
	\end{split}
\end{equation}
Where $\beta$ is the infectious rate, $\gamma$ is the exposed rate, $\delta$ is the recovery rate, and $N=S+I+E+R$ is the number of the total population.

The SEIR models are simple and only use the infectious rate, exposed rate, recovered rate to describe the development trend of infectious diseases. However, the infection situation in the real world is complicated. Some existing studies based on SIR or SEIR models propose modeling improvements to fit the different population trends \cite{10Peng, 18Wei, 18Choi}. These works often use a complex improved SEIR model with multiple parameters to analyze a specific area to achieve high-precision fitting performance. However, due to each region's particularities (different population conditions, social distance, isolation measures, etc.), it is difficult to directly migrate these models to other areas. Developing a different model for a specific region is very time-consuming, but precise modeling can help estimate the spread of COVID-19 diseases. Besides, because the infection situations between different areas are also very inconsistent (some countries have just begun to spread while the infection cases in others have disappeared), different areas may use different models to analyze. Therefore, simple models with high fitting accuracy that are more generally adapted to various countries or regions and more easily migrated to another area are of great significance in the modeling of COVID-19.

Machine learning has achieved remarkable results in solving many complex data-driven problems such as medical data prediction \cite{TBME_pred1, TBME_pred2, TBME_pred3, TBME_pred4}. Thus, it is also used to model the COVID-19 data \cite{AI_covid, 11Yang, TBME_COVID}, offering great data fitting capabilities in early data modeling. However, the forecasts under different development situations in the middle and late periods have not been explored. Furthermore, these machine learning models' black-box mechanism allows their users to obtain the prediction results directly. However, it is unclear what the reason is the prediction. Thus, these models have low interpretability, and their prediction process cannot be analyzed and quantified.

Due to the limitations of existing approaches, we propose a deep-learning model called deep dynamic epidemiological (DDE), which combines neural ordinary differential equations (Neural ODE) \cite{12Chen} and epidemiological equations in COVID-19 data fitting. Because of different countries and regions' characteristics, we propose four variant models based on SIR and SEIR models, namely SIRD, SEIRD, SMCRD, and SEMCRD. These four models consider the population group division's situation and can deal with different data sources (for example, some countries' published information does not give the specific number of mild and severe cases). Secondly, we developed the DDE implemented on these four variants to solve the equations. In our DDE model, we based on the Neural ODE, a novel algorithm, which can use numerical solutions of ordinary differential equations to build networks and complete data fitting and modeling. Also, we design an additional neural network under Neural ODE solving to fit the effect function to reflect better the impact of the diversity of regional and national intervention policies on the infection rate. Therefore, based on understanding the parameters, including infection rate, recovery rate, mortality rate, and so on, involved in the SEIR model, it can obtain outstanding solution accuracy. The DDE is used to solve the problem, which can better consider the impact of different isolation policies and realize the possibility of designing a universal model. Specifically, the contributions of this paper are summarised as follows.
\begin{enumerate}
	\item We propose the DDE method that can easily integrate effect stimulation networks and Neural ODE to solve SEIR-like equations. It provides infection, mortality, and recovery rate through network training and can visualize the change of those rates. Moreover, DDE has achieved Pearson correlation coefficient above 0.98 in all data.
	\item  We design four variants: SIRD, SEIRD, SMCRD, and SEMCRD, which can adapt to regions and countries' diversity. Further, we implement the DDEs based on the four variants are more precise than traditional parameter estimation methods.
	\item  We compare the performance of DDE method and other machine-learning and deep-learning models. The DDEs gain average Pearson coefficients higher than 0.85. Further, we illustrate the parameter trend which reflect real-world meaning.
\end{enumerate}

The rest of this paper is organized as follows. Section II introduces related works of modeling COVID-19 data. Section III offers a brief introduction to the four variants we proposed: SMCRD and SEMCRD models. Also, we introduces our DDE methods in Section III. In Section IV, we present the experiment to illustrate the DDE-SIRD, DDE-SMCRD, DDE-SEIRD, and DDE-SEMCRD models' fitting performance. Moreover, the comparisons of our DDE and parameter estimation methods and machine-learning or deep-learning methods are shown in Section IV. Finally, Section V concludes the work of this paper.

\begin{figure*}[ht]
	\begin{center}
		\includegraphics[width=6.5in]{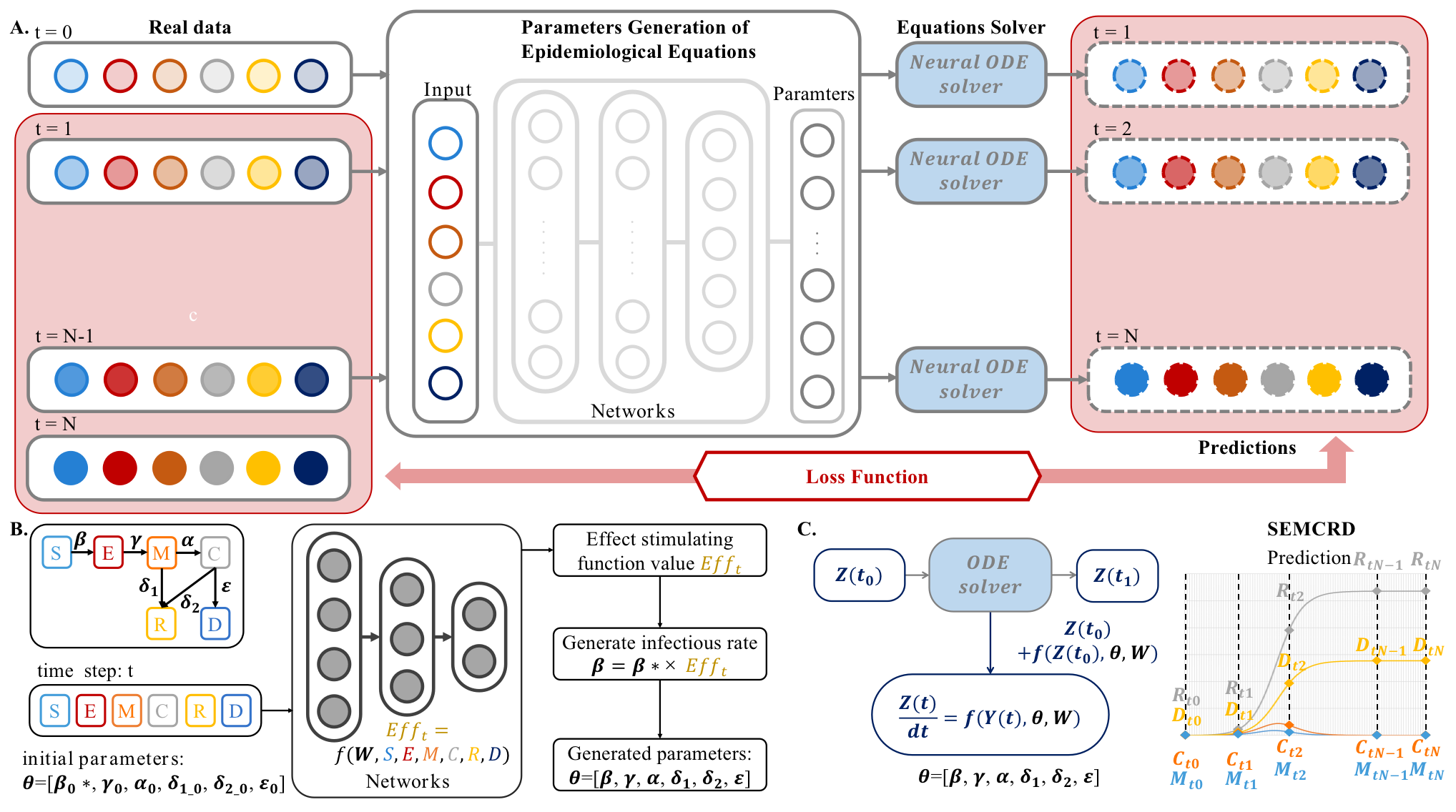}
		\caption{The schematic diagram shows the system structure of the DDE model. \textbf{A.} The method architecture shows in Part A. The real data are represented by solid frames, and the predictions are shown by dotted frames. The system consists of four parts: First, we input the value of the equations at the current moment into the parameter generation network to obtain the corresponding parameters. After that, the generated parameters and current values are used to obtain the function value at the next moment through the Neural ODE solver. Then, repeat the process several times until the solution obtains the predicted value. Finally, we calculate the loss function between the predicted value and the true value to obtain the best prediction through gradient descent. \textbf{B.} The process figure shows the details of generating parameters. \textbf{C.}The schematic plot illustrates the process of solving the epidemiological equations..}
		\label{fig:structure}
	\end{center}
\end{figure*}

\section{Related Work}
SIR and SEIR models are commonly used in epidemiological analysis. In the SEIR model, the country's population is divided into four parts: susceptible people S, exposed people $E$, infected patients $I$, and recovered group $R$, and their relative growths are based on a set of coupled ordinary differential equations. The SIR model is not considered the exposed group E. In the SARS and MARS, which once caused the global epidemic, many existing studies are based on the SEIR model to model and analyze infectious diseases \cite{5Fang, 6Saito, 7Smirnova}. Moreover, in COVID-19 studies, many works focus on improved SEIR modeling based on region or cities. In the study of L. Peng et al. \cite{10Peng}, they proposed a generalized SEIR model to analyze the COVID-19 epidemic in China. Additionally, \cite{18Choi} Choi's work introduced an improved SEIR model called the SEIAQIm model based on Korea's data. These improved SEIR models are aimed at a specific country or region and have quite complex parameter designs.

Furthermore, statistical models are widely used in specific aspect modeling of real-world data. For example, in Kraemer's work \cite{9Kraemer}, the author used Generalized linear models (GLM) to analyze the effect of human mobility and control measures in early Wuhan, China. Furthermore, in \cite{8Chinazzi}, a global epidemic and mobility model (GLEAM) is proposed to analyze the effect of travel restriction on the spread of COVID-19 in the world. This work used experienced parameters obtained in SARS or MARS coronavirus epidemiology instead of independently from COVID-19 data. Moreover, work \cite{20Kucharski} focuses on the impact of cases exported from Wuhan on other regions using a stochastic transmission dynamic model. The widely used statistical models have brought valuable solutions to the analysis of some specific aspects, but the models are diverse, and the designs are complex.

Some studies on the COVID-19 data trend prediction issue have also introduced machine learning attempts due to the SEIR models and statistical methods' deficiencies. Z. Yang's work \cite{11Yang} integrated the population migration data before and after January 23 and the latest COVID-19 epidemiological data into the SEIR model to derive the epidemic curve. Also, it used artificial intelligence (AI) methods trained on 2003 SARS data to predict the epidemic. Besides, F. Rustam et al.\cite{AI_covid} proposed to use some machine learning model including linear regression (LR), the least absolute shrinkage and selection operator (LASSO), support vector machine (SVM), and exponential smoothing (ES) to predict the population of infection, recovery, and death. These attempts either used previous infectious disease data for training, or the models used were insufficient in terms of interpretability. Moreover, in work \cite{Dandekar2020}, researchers tried to used the Neural ODE method to stimulate data change in the infection and recovery group. However, this method only explores the SIR model's improved performance and the fit of real data. This method has only been experimentally studied on the early development of COVID-19 (before April) and has not compared model performance with other methods. Furthermore, the way does not analyze the data fitting situation of different size regions (countries, regions, and cities). Therefore, these factors also hinder the application of the advanced Neural ODE method in practice. 

\section{Proposed Method}
In this section, we propose a data-driven model for real-world infection data fitting. First, we introduce four variants based on SEIR and SIR models for better fitting of COVID-19 data. Next, the DDE method is illustrated for dynamic parameter estimation. In our DDE, multi-layer neural networks are designed for stimulating intervention influence function to assess the effect of quarantine policies in different countries. Furthermore, Fig \ref{fig:structure} illustrates the computing process in our DDE method for dynamic parameter estimation.

\subsection{Four variants: SIRD, SEIRD, SMCRD, and SEMCRD}
The classic SEIR model has been employed in countless prior studies \cite{5Fang, 6Saito, 7Smirnova}. In research about COVID-19, the SEIR model is also prevalent \cite{13Read, 14Tang, 15Wu}. According to the characteristics of the COVID-19, we consider the following aspects of improving the initial SEIR and SIR models:

\begin{figure}[ht]
	\begin{center}
		\includegraphics[width=3.5in]{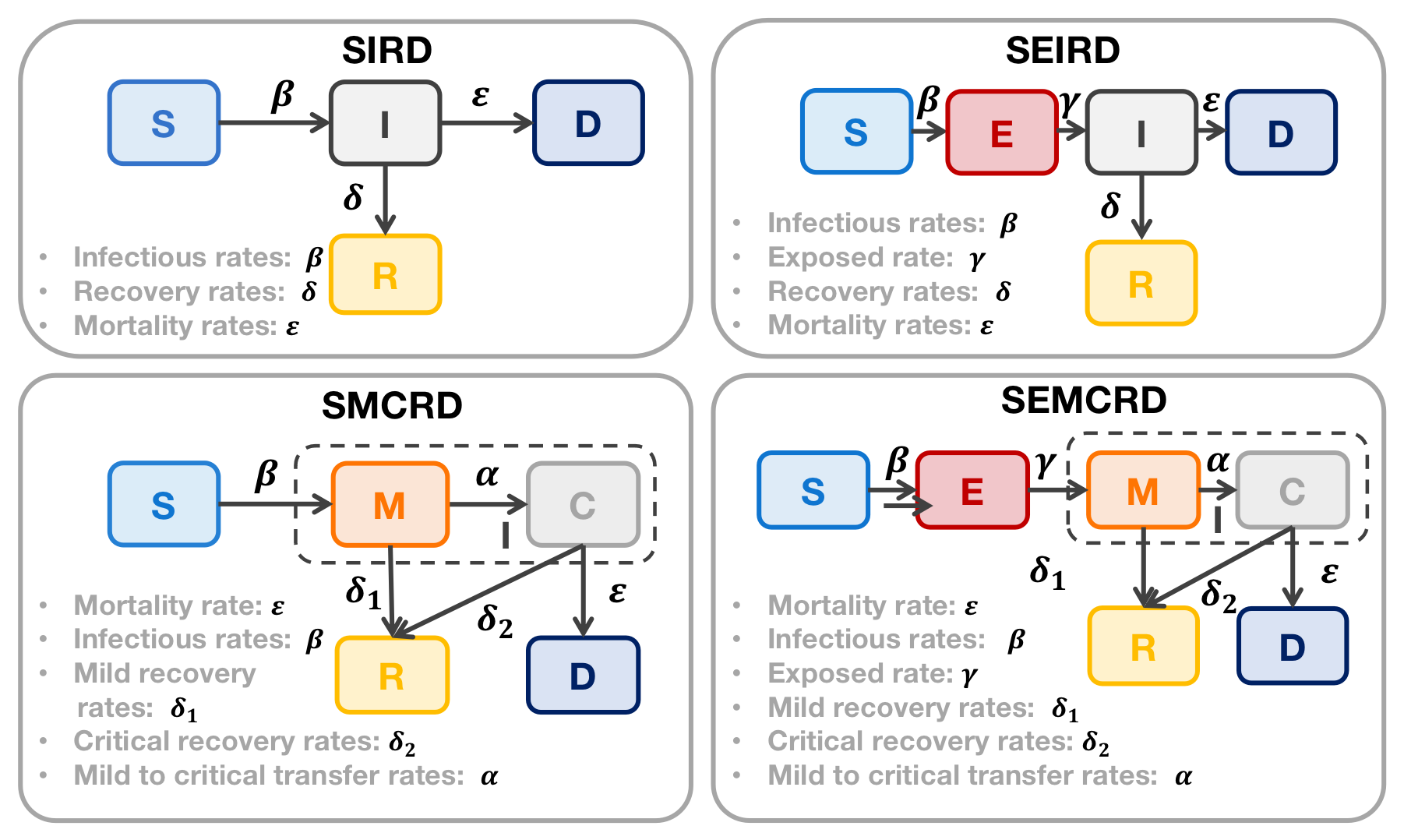}
		\caption{The architectures display the four variants based on SEIR and SIR models: SIRD, SMCRD, SEIRD, SEMCRD. In the SIRD model, the infection rate $\beta$, recovery rate $\delta$, and mortality rate $\varepsilon$ are considered. Furthermore, the SEIRD model takes the expose group $E$ into account and assumes the exposed rate as $\gamma$. The SMCRD and SEMCRD models divide the infection group into mild patient $M$ and critical patient $C$. Thus, the probability of transferring mild patients to severe patients is $\alpha$, the recovery rate of the mild patient is $\delta_1$, and the recovery rate of the critical patient is $\delta_2$.}
		\label{fig:SIRD_stru}
	\end{center}
\end{figure}

\begin{enumerate}[]
	\item  We make a new way to classify the rehabilitation population (including death and disease rehabilitation). The original recovery group is divided into the death population and recovery (disease rehabilitation) population. We propose two parameters (mortality and recovery rate) to better respond to different countries' death and recovery situations due to different medical conditions. 
	\item We accurately define the infected population as the mild population (including asymptomatic infected persons, self-recovering mild patients) and critical patients (including severe patients who need to be admitted to hospital). Further, we consider three parameters: the mild infectious rate, the transition rate from mild to critical,  the mild recovery rate, and the critical recovery rate, helping us better simulate real infections.
\end{enumerate}

Therefore, we implement four variants: SIRD, SEIRD, SMCRD, and SEMCRD.  Fig. \ref{fig:SIRD_stru} shows the model structures of the four models. As shown in Fig. \ref{fig:SIRD_stru}, compared with the SIR and SEIR models, the SIRD and SEIRD models add the consideration of the death population, which can help understand the death growth caused by diseases. On the other hand, SMCRD and SECRD are based on the SIRD and SEIRD models and further divide the infection group I into mild and critical cases. The SMCRD and SEMCRD models are in line with the characteristics of the COVID-19 disease and can be better modeled in real-world data.	

The equation of these models can be seen as follow:
\begin{equation}
	\begin{split}
		\frac{d\bm{Z}(t)}{dt}=\mathscr{F}(\bm{Z}(t), t, \theta), \text{ with } \bm {Z}(t_0)=\bm{Z}_0
	\end{split}
\end{equation}
where $t \in \{t_0, ..., t_i, ..., t_T\}$ ($t_0$ stands for the initial day, and $t_i$ represents the $i^{th}$ day from $t_0$), $\bm{Z}(t) \in \mathbb{R}^D$, $\bm{Z}_0 = [S_0, E_0, I_0, R_0, D_0]$ for SEIRD model, $\bm{Z}_0 = [S_0, I_0, R_0, D_0]$ for SIRD model, $\bm{Z}_0 = [S_0, E_0, M_0, C_0, R_0, D_0]$ for SEMCRD model, and $\bm{Z}_0 = [S_0, M_0, C_0, R_0, D_0]$ for SMCRD model, $\bm{Z}_0$ is the function value at time $t_0$. The $\mathscr{F}(\cdot)$ is a known and continuous function with parameter $\theta$, and $\bm{Z}(t)$ is the unknown function that must be approximated. For SIRD model, the parameter set $\theta$ includes $\beta, \delta, \varepsilon$. The SEIRD model's parameter set $\theta$ includes $\beta,\gamma, \delta, \varepsilon$. Further, for SMCRD model, $\theta$ includes $\beta, \delta_1, \delta_2, \alpha, \varepsilon$. In SEMCRD model, $\theta$ includes $\beta, \gamma, \delta_1, \delta_2, \alpha, \varepsilon$. As Fig. \ref{fig:SIRD_stru} shows, the epidemiological equations for SEMCRD model can be seen in the bottom right part, and the equation set in SMCRD model is in the bottom left part. The difference between these two models is that the SMCRD model does not consider the exposed group $E$.

From a computational point of view, knowing that $\bm{Z}(t_0) = \bm{Z}_0$, you can calculate the value of $\bm{Z}(t_i) = \bm{Z}_i$ in any step $t_i$ by performing piecewise integration from previously known points:
\begin{equation}
	\begin{split}
		\bm{Z}_i = \bm{Z}_{i-1} + \int_{t_{i-1}}^{t_i} \mathscr{F}(\bm{Z}_{i-1},t_{i-1},\theta) \text{ with } i \in \{1, ..., T\}
	\end{split}
\end{equation}
When $\Delta\tau = t_i-t_{i-1}$ is small enough, we can get the approximation result:
\begin{equation}
	\begin{split}
		\bm{Z}_i = \bm{Z}_{i-1} + \Delta\tau \cdot \mathscr{F}(\bm{Z}_{i-1},t_{i-1},\theta)
	\end{split}
\end{equation}
Thus, in each time step $t_i$, the value of function $\bm{Z}_i$ can be obtained by deduction of the function value $\bm{Z}_{i-1}$ at the previous moment $t_{i-1}$.

\subsection{DDE Method applying in SMCRD and SEMCRD models}
\begin{figure}[ht]
	\begin{center}
		\includegraphics[width=3.5in]{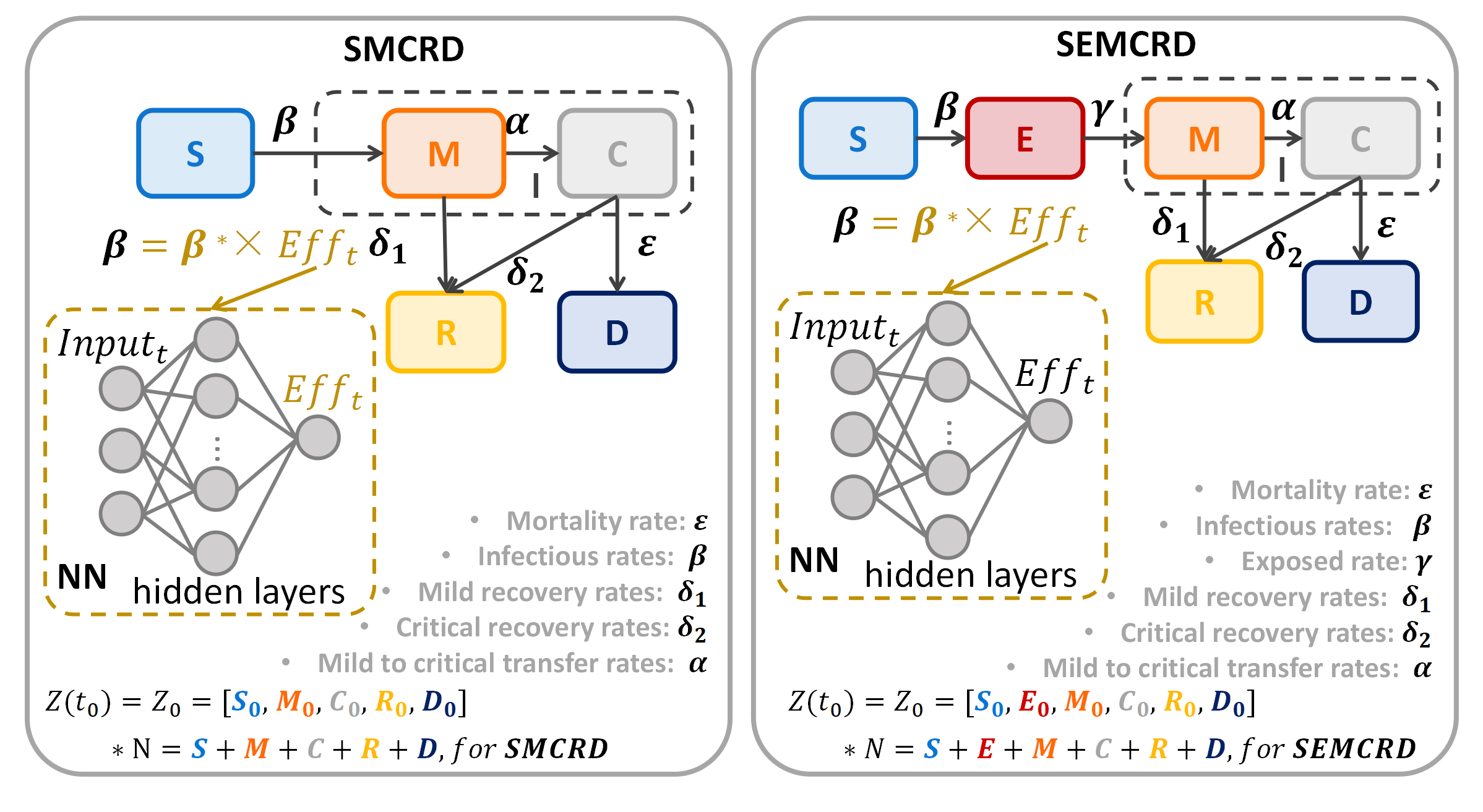}
		\caption{The DDE method implements in SMCRD and SEMCRD models.}
		\label{fig:DDE}
	\end{center}
\end{figure}

In real-world data, the infection condition is always influenced by social distancing, quarantine measurement, people's compliance. In recent SEIR model studies, the effect function based on the factors mentioned above is considered. In our DDE model, we also propose a way to stimulate the effect function: we design multi-layer neural networks to fit the effect function's value and use it as a dynamic parameter involved in the solution process of Neural ODE. Fig \ref{fig:DDE} illustrates how to use multi-layer neural networks in parameter designing. The estimation of the designed parameters bases on the Neural ODE method.  

Neural ODE \cite{12Chen} method is a novel technique that used reverse-mode differentiation (also known as backpropagation) to solve the ODE function. In our work, the ODE equation sets in SEMCRD and SMCRD models are shown above. Using the Neural ODE method in our task can make full use of neural networks' fitting performance within the framework of epidemiological equations. Our model uses infection rate, recovery rate, mortality rate, and so on, which have real-world meanings as parameters. Also, we use neural networks to perform fitting to achieve high precision and high interpretability.

\begin{figure}[ht]
	\begin{center}
		\includegraphics[width=3.5in]{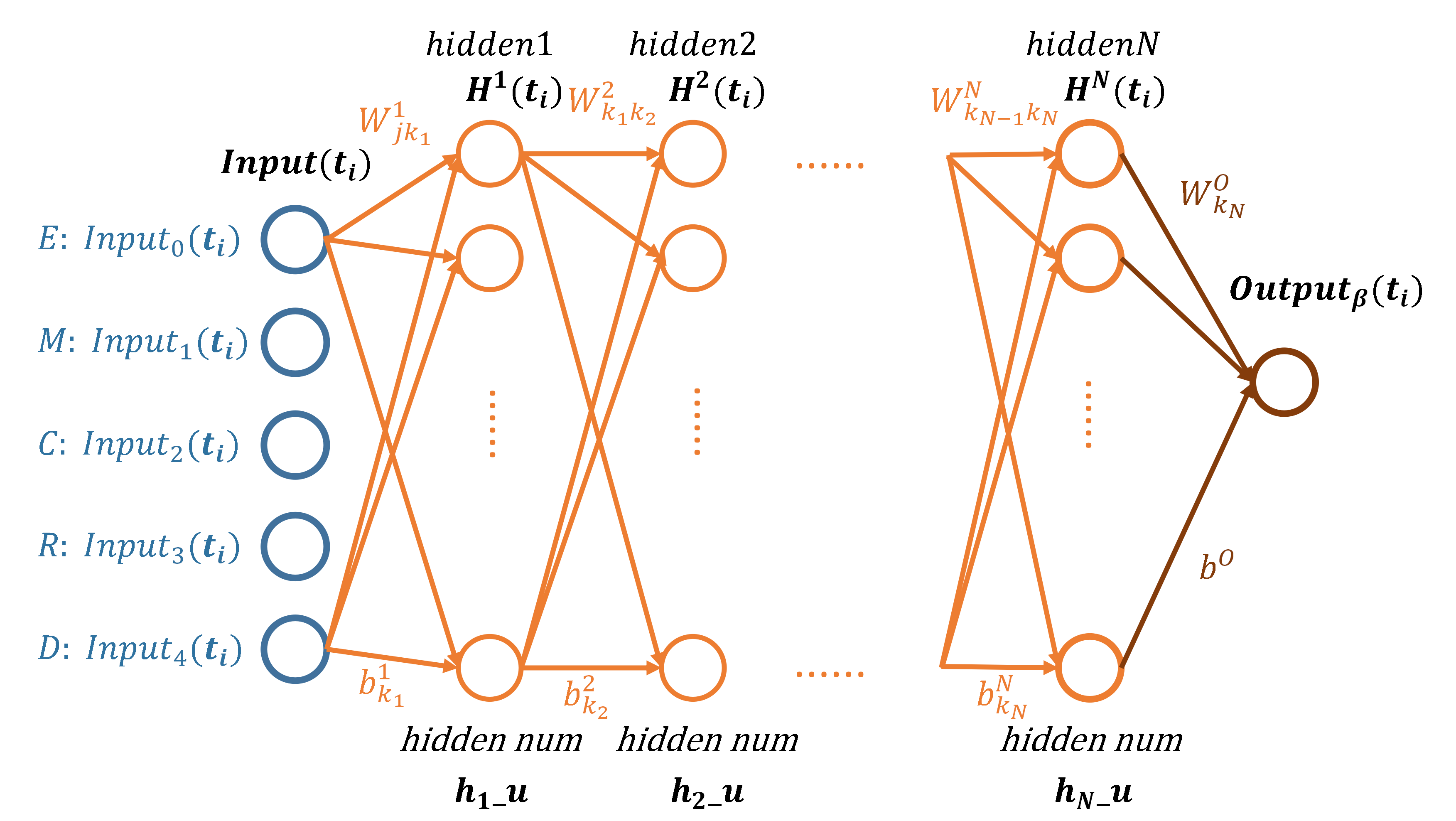}
		\caption{The structure figure shows the multi-layer neural networks which are used to fit the effect function $Eff_t$ on infection rate $\beta$.}
		\label{fig:nn}
	\end{center}
\end{figure}

Neural networks are known for their strong fitting ability. Also, because of Neural ODE's characteristics, we could develop multi-layer neural networks as an influence function value stimulator that reflects the influence on infection rate.  In the previous part, we propose four variants based on SEIR or SIR model and give the design of parameters involved in SIRD, SEIRD, SMCRD, and SEMCRD. Further, in real data fitting, different countries and regions often adopt different intervention isolation measures and policies, and people respond to policies differently. The impact of these problems on the infection rate is also different. For example, in some countries, very strict movement bans have been implemented, which can highly impact the infection situation. The sharp drop in the number of contacts will also slow down the increase in the number of people infected with COVID-19. In countries that have not adopted control measures, the infection situation's impact may be small, and the number of infections will continue to increase. Thus, we develop an effect stimulation function to model the various change of infection situations. The parameter set $\theta$ is changed:
\begin{equation}
	\begin{split}
		\theta=[\beta*, W, \delta, \varepsilon]\text{ , for SIRD}\\
		\theta=[\beta*, W, \gamma, \delta, \varepsilon]\text{ , for SEIRD}\\
		\theta=[\beta*, W, \delta_1, \delta_2, \alpha, \varepsilon]\text{ , for SMCRD}\\
		\theta=[\beta*, W, \gamma, \delta_1, \delta_2, \alpha, \varepsilon]\text{ , for SEMCRD}\\
	\end{split}
	\label{equ:theta}
\end{equation}
where the $\beta*$ is the infection rate which could be influenced by effect function $Eff_t$. The effect function $Eff_t$ use the multi-layer neural networks to fit. The $W$ is weights in the multi-layer neural networks. The structure of the neural networks is shown in Fig. \ref{fig:nn}.

According to Fig. \ref{fig:nn}, we can obtain the output of the first hidden layer of the neural network as follow:
\begin{equation}
	\begin{split}
		H_{k_1}^1(t_i) = a(\sum_{j=1}^{input_u}W_{j \cdot {k_1}}^1 \cdot Input_{j}(t_i)+b_{k_1}^1)\\
	\end{split}
\end{equation}
where $H_{k_1}^1(t_i)$ is the ${k_1}^{th}$ of unit of the first hidden layer output in time step $t_i$, the input tensor $Input(t_i)$ =[$Input_0(t_i)$,  $Input_1(t_i)$, $Input_2(t_i)$, $Input_3(t_i)$, $Input_4(t_i)$] for SEMCRD ($Input_0$ is $E_{t_i}$, $Input_1(t_i)$ is $M_{t_i}$, $Input_2(t_i)$ is $C_{t_i}$, $Input_3(t_i)$ is $R_{t_i}$, and $Input_4(t_i)$ is $D_{t_i}$) and $Input(t)$ =[$Input_0(t_i)$, $Input_1(t_i)$, $Input_2(t_i)$, $Input_3(t_i)$] for SMCRD ($Input_0(t_i)$ is $M_{t_i}$, $Input_1(t_i)$ is $C_{t_i}$, $Input_2(t_i)$ is $R_{t_i}$, and $Input_3(t_i)$ is $D_{t_i}$), the $input_u$ is the number of $Input(t_i)$, it is 5 for SEMCRD model or is 4 for SMCRD model. The $W_{j \cdot {k_1}}^1$ represent the weight of input layer and the first hidden layer in the connection weight of the $j^{th}$ input unit and the ${k_1}^{th}$ of the first hidden unit. The $b_{k_1}^1$ is the bias of the ${k_1}^{th}$ unit of the first hidden layer, and the $a(\cdot) = ELU(\cdot)$ is the activation function of hidden layer.

The output of the $2^{rd}$ hidden layer to the $N^{th}$ hidden layer also can be deduced in the following:
\begin{equation}
	\begin{split}
		H_{k_2}^2(t_i) = a(\sum_{k_{1}=1}^{h_{1}\_u}W_{k_{1} \cdot {k_2}}^2 \cdot H_{k_{1}}^{1}(t_i)+b_{k_{2}}^2)\\
		......\\
		H_{k_N}^N(t_i) = a(\sum_{k_{N-1}=1}^{h_{N-1}\_u}W_{k_{N-1} \cdot {k_N}}^N \cdot H_{k_{N-1}}^{N-1}(t_i)+b_{k_{N}}^N)\\
	\end{split}
\end{equation}
where $a(\cdot)$ is the activation function, and all hidden layers use the same activation function, $h_{N-1}\_u$ is the number of hidden units in the ${N-1}^{th}$ hidden layer, $b_{k_{N}}^{N}$ is the bias of the ${k_N}^{th}$ unit of the $N^{th}$ hidden layer, $W_{k_{N-1} \cdot k_N}^N$ represents the parameters between the $N-1^{th}$ hidden layer and $N^{th}$ hidden layer. And, the $H_{k_{N-1}}^{N-1}(t_i)$ is the ${k_{N-1}}^{th}$ units of the ${N-1}^{th}$ hidden layer in time step $t_i$, and $H_{k_{N}}^{N}(t_i)$ is the ${k_{N}}^{th}$ units of the ${N}^{th}$ hidden layer in time step $t_i$. Similarly, replace N with 2 to get the mean of $H_{k_2}^2(t_i)$, $h_{1}\_u$, $W_{k_{1} \cdot {k_2}}^2$, $b^2$, and $H_{k_{1}}^{1}(t_i)$.

\begin{figure*}[ht]
	\begin{center}
		\includegraphics[width=6.2in]{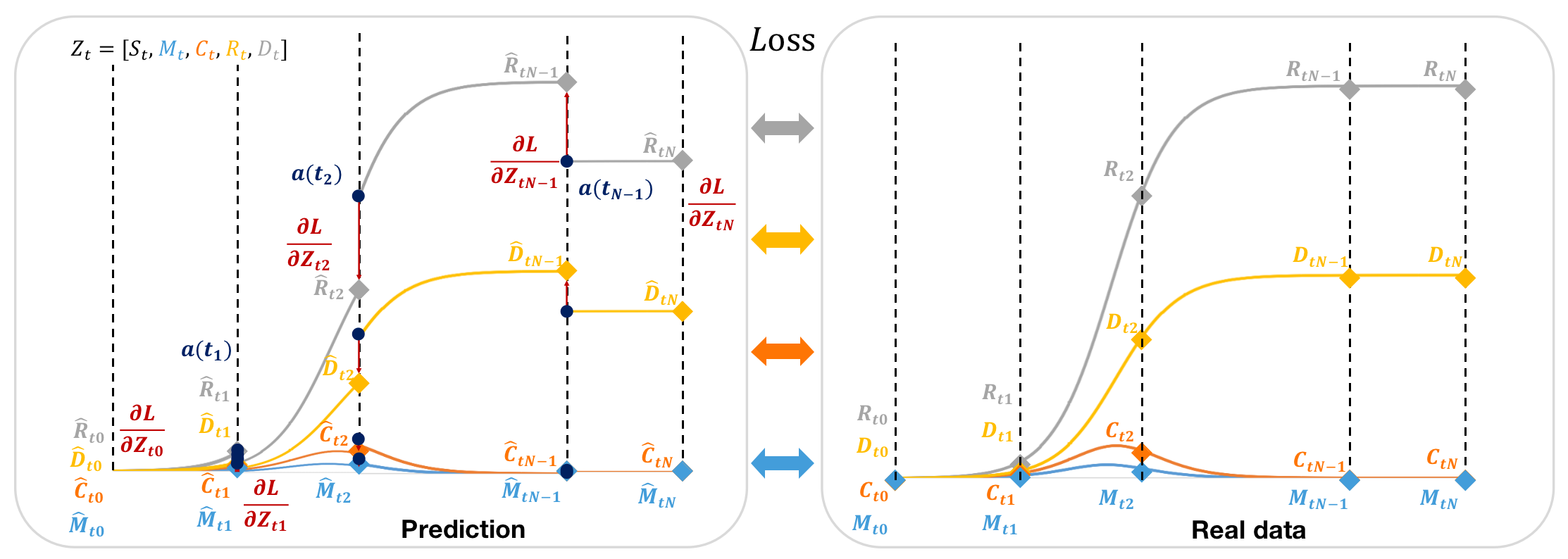}
		\caption{The computing process of loss value $L$. If the loss directly depends on the state of multiple observation times, the adjoint state $a(t)$ must be updated in the direction of the partial derivative of the loss of each observation. In addition to describing the initial dynamic system process, the state $a(t)$ also describes the derivative state of each point in the reverse process through the chain rule.}
		\label{fig:process}
	\end{center}
\end{figure*}

Thus, we can calculate the output of this network:
\begin{equation}
	\begin{split}
		Output_{\beta}(t_i) = (1+e^{-\sum_{k_N=1}^{h_{N\_u}}W_{k_N}^O \cdot H_{k_N}^N(t_i)+b^O})^{-1}
	\end{split}
\end{equation}
where $Output_{\beta}(t_i)$ in the $i^{th}$ time step $t_i$ as the output of $N$-layer NN, $h_{N}\_u$ is the number of hidden units in the $N^{th}$ hidden layer, $W^O$ represents the parameters between the $N^{th}$ hidden layer and the output layer and the $W_{k_N}^O$ stands for the ${k_N}^{th}$ of the weight matrix $W^O$, $H^N(t_i)$ is the output of the $N^{th}$ hidden layer in time step $t_i$ and the $H_{k_N}^N(t_i)$ is the output of the ${k_N}^{th}$ of the hidden layer output, and the $b^O$ stands for the bias of the output layer.
Moreover, we can present a simper form of $Output_{\beta}(t_i)$ with weight of $N$ hidden layer is $W = \{W^1, W^2, ..., W^N\}$, and $W^O$ is the weight of output layer as:
\begin{equation}
	\begin{split}
		Output_{\beta}(t_i) = a_O(W^O(a(W^{N}...\\
		a(W^1 \cdot Input(t_i)+b^1)...)+b^N)+b^O)\\
	\end{split}
\end{equation}
where the activation function of output layer is $a_O(\cdot)$, and the activation function of hidden layers is $a(\cdot)$. The bias $\{b^1, ...,b^N\}$ is of the $1^{th}$ to $N^{th}$ of hidden layers.

The output of the neural networks represent the effect function $Eff_t$. To solve the ODE equations $\bm{Z}(t)$, we need to use ODE solver. Efficient and accurate ODE solvers has been used during 120 years, and can guarantees about the growth of approximation error, monitor the level of error, and adapt their evaluation strategy on the fly to achieve the requested level of accuracy \cite{12Chen}. In specific time $t_i$, the function $\bm{Z}_i$ can be obtained into the form:
\begin{equation}
	\begin{split} \label{ode}
		\bm{Z}_i =ODESlover(\bm{Z}_{i-1}, \mathscr{F}(\bm{Z}_{i-1},t_{i-1},\theta) , t_{i-1}, \theta), \\\text{ with } \bm{Z}(t_0)=\bm{Z}_0
	\end{split}
\end{equation}
According to the equation (\ref{ode}), each $\bm{Z}_i$ can be calculated. Different models have various parameter set $\theta$. The specific content of $\theta$ has been given in Equ. \ref{equ:theta}. 

Further, we design the loss function. The loss function $L$ with multi-layer neural networks attempt to minimize the value of loss function by adjust parameters including $\beta*, W, \gamma, \delta_1, \delta_2, \alpha, \varepsilon$, and weight of the multi-layer NN for SEMCRD model, and  $\beta*, W, \delta_1, \delta_2, \alpha, \varepsilon$, and weight of the multi-layer NN for SMCRD model. The simper loss function is:
\begin{equation}
	\begin{split}
		L(X, \hat{X}) =\frac{1}{3} \sum_{t=0}^T \min_{\theta} \{ (log(X_t)-log(\hat{X_t}))^2\}\\
		= \frac{1}{3} \sum_{t=0}^T\min_{\theta, W}\{||log(I_t)-log(\hat{M}_t+\hat{C}_t)||^2\\+||log(R_t)-log(\hat{R}_t))||^2+||log(D_t)-log(\hat{D}_t)||^2\}
	\end{split}
\end{equation}
Where $X = \{X_0, X_1, ..., X_T\}$ represent real data considered in cost function in each time step and $\hat{X} = \{\hat{X}_0, \hat{X}_1, ..., \hat{X}_T\}$ stands for predicted data per day. In the $i^{th}$ of time $t_i$, the real data used in calculate loss is $X_i = [I_i, R_i, D_i]$, and the predicted data is $\hat{X}_i = [\hat{M}_i+\hat{C}_i, \hat{R}_i, \hat{D}_i]$ ($I_i$, $R_i$, and $D_i$ represent the number of real infection, recovery, death group, relatively, and $\hat{M}_i$, $\hat{C}_i$, $\hat{R}_i$, and $\hat{D}_i$ stand for the number of predicted mild patients, critical patients, recovery, death group.).

To minimize $L$, understanding how the gradient of the loss depends on equation set $\bm{Z}(t)$. According to the chain rules, we can know that:
\begin{equation}
	\begin{split}
		\frac{d\bm{a}(t)}{dt} = -\bm{a}(t)^T\frac{\partial f(\bm  Z(t),t,\theta)}{\partial \bm{Z}}\\
		\text{where,} \bm{a}(t)=\frac{\partial L}{\partial \bm{Z}(t)}
	\end{split}
\end{equation}
The $\frac{\partial L}{\partial \bm{Z}(t_0)}$ can be computed by the $ODESolver$ mention above. The $ODESolver$ must start from the initial value of $\frac{\partial L}{\partial \bm{Z}(t_T)}$ and run backwards. A complicated situation is that solving this ODE requires knowing the value of $\bm{Z}(t)$ along its entire trajectory. However, we can simply recalculate $\bm{Z}(t)$ from its final value $\bm{Z}(t_T) = \bm{Z}_T$ and work backwards with time.

The gradients of loss function $L$ depends on both $\bm{Z}(t)$ and $\bm{a}(t)$:
\begin{equation}
	\begin{split}
		\frac{dL}{d\theta} = - \int_{t_T}^{t_0}\bm{a}(t)^T\frac{\partial f(\bm{Z}(t),t,\theta)}{\partial \theta}dt
	\end{split}
\end{equation}

The $\bm{a}(t)^T\frac{\partial \mathscr{F}}{\partial \bm{Z}}$, $\bm{a}(t)^T\frac{\partial \mathscr{F}}{\partial \theta}$ and $\mathscr{F}$ can be obtained through automatic differentiation of computing vector-Jacobian products. Finally, all gradients including $\frac{\partial L}{\partial \bm{Z}(t_0)}$ and $\frac{\partial L}{\partial \theta}$ can be calculated at once by calling the ODE solver.
The computing progress of the gradient of loss function $L$ shows in Fig. \ref{fig:process}. From the steps, we first define the initial state, and then design the dynamics system. After that, we compute vector-Jacobian products to obtain $\bm{a}(t)^T\frac{\partial \mathscr{F}}{\partial \bm{Z}}$, $\bm{a}(t)^T\frac{\partial \mathscr{F}}{\partial \theta}$ and $\mathscr{F}$. In the end, the $ODESolver$ are used for solve the reverse-time SMCRD and SEMCRD equations, and gain the gradients.

\begin{table*}
	\caption{Mean square error (MSE) of SIR, SEIR, SIRD, SEIRD, SMCRD, SEMCRD models in different optimization methods, including Neural ODE(Adam), minimizing algorithms(Nelder-Mead, Powell, BFGS, Truncated Newton Conjugate-Gradient (TNC)). The unit of MSE is ten thousand people.}
	\label{tab:optimizedr}
	\centering
	\begin{threeparttable}
		\begin{tabular}{c|c|c|c|c|c|c|c|c|c|c|c}
			\hline
			\multirow{3}*{Models} & Parameter & \multicolumn{6}{c|}{Courtries} & \multicolumn{4}{c}{Regions}\\
			\cline{3-12}
			& Estimation & \multicolumn{2}{c|}{CO} &  \multicolumn{2}{c|}{USA} & \multicolumn{2}{c|}{ZA} & \multicolumn{2}{c|}{CN-WH} & \multicolumn{2}{c}{Italy-PD}\\
			\cline{3-12}
			& Method & MSE & Pearson & MSE & Pearson & MSE & Pearson & MES & Pearson & MSE & Pearson\\
			\hline
			\multirow{6}*{SIR \cite{5Fang}}& Neural ODE \cite{12Chen} & 19.317  & 91.95$\%$ & 16550.121  & 72.04$\%$ & 183.378  & 82.92$\%$ & 2.407  & 34.71$\%$ & 4.697  & 49.77$\%$\\
			& Nelder-Mead \cite{Nelder_Mead} & 40.253  & 84.53$\%$ & 2979.545  & 86.98$\%$ & 166.426  & 86.98$\%$ & 5.875  & 29.60$\%$ & 11.609  & 46.82$\%$\\
			& Powell \cite{Powell} & 30.002  & 91.28$\%$ & 2093.261  & 88.19$\%$ & 821.494  & 89.45$\%$ & 0.851  & 48.41$\%$ & 3.714  & 55.91$\%$\\
			& BFGS \cite{BFGS} & 28.419  & 88.23$\%$ & 2094.101  & 80.29$\%$ & 196.961  & 85.82$\%$ & 1.125  & 56.80$\%$ & 8.182  & 47.27$\%$\\
			& TNC \cite{TNC} & 27.632  & 80.80$\%$ & 2703.972  & 86.92$\%$ & 54.257  & 91.69$\%$ & 1.099  & 47.28$\%$ & 5.102  & 39.98$\%$\\
			& DDE & $\bm{0.210}$  & $\bm{99.17\%}$ & $\bm{52.799}$  & $\bm{97.14\%}$ & $\bm{1.143}$  & $\bm{97.93\%}$ & $\bm{0.140}$  & $\bm{98.33\%}$ & $\bm{0.068}$  & $\bm{95.18\%}$\\
			\hline
			\multirow{6}*{SEIR \cite{6Saito}} & Neural ODE \cite{12Chen} & 23.231  & 90.64$\%$ & 10523.219  & 78.31$\%$ & 38.832  & 90.64$\%$ & 0.967  & 90.77$\%$ & 6.812  & 59.33$\%$\\
			& Nelder-Mead \cite{Nelder_Mead} & 22.111  & 90.18$\%$ & 2172.293  & 82.98$\%$ & 195.785  & 88.69$\%$ & 5.677  & 22.02$\%$ & 8.903  & 32.67$\%$\\
			& Powell \cite{Powell} & 51.602  & 84.88$\%$ & 2686.344  & 82.63$\%$ & 190.284  & 87.94$\%$ & 1.950  & 45.99$\%$ & 4.922  & 65.87$\%$\\
			& BFGS \cite{BFGS} & 33.297  & 63.17$\%$ & 2767.012  & 84.23$\%$ & 78.357  & 26.39$\%$ & 8.627  & 55.88$\%$ & 24.941  & 47.55$\%$\\
			& TNC \cite{TNC} & 30.671  & 57.44$\%$ & 2882.301  & 80.28$\%$ & 36.361  & 91.25$\%$ & 6.777  & 69.52$\%$ & 3.573  & 35.62$\%$\\
			& DDE & $\bm{0.192}$  & $\bm{99.77\%}$ & $\bm{56.901}$  & $\bm{98.75\%}$ & $\bm{0.967}$  & $\bm{99.01\%}$ & $\bm{0.250}$  & $\bm{98.03\%}$ & $\bm{0.034}$  & $\bm{94.47\%}$\\
			\hline
			\multirow{6}*{SIRD} & Neural ODE \cite{12Chen}& 29.148  & 90.82$\%$ & 10190.655  & 70.90$\%$ & 137.386  & 85.09$\%$ & 2.910  & 31.79$\%$ & 4.063  & 42.98$\%$\\
			& Nelder-Mead \cite{Nelder_Mead}& 59.437  & 89.95$\%$ & 2722.980  & 82.56$\%$ & 177.995  & 86.01$\%$ & 5.303  & 29.72$\%$ & 8.003  & 40.54$\%$\\
			& Powell \cite{Powell}& 23.496  & 93.31$\%$ & 2105.093  & 84.57$\%$ & 84.720  & 89.22$\%$ & 0.975  & 51.43$\%$ & 3.714  & 45.74$\%$\\
			& BFGS \cite{BFGS}& 57.518  & 90.02$\%$ & 2693.094  & 82.00$\%$ & 176.266  & 85.89$\%$ & 1.067  & 23.86$\%$ & 8.582  & 40.10$\%$\\
			& TNC \cite{TNC}& 29.590  & 92.58$\%$ & 2218.704  & 85.08$\%$ & 59.395  & 90.61$\%$ & 1.128  & 44.17$\%$ & 3.179  & 46.93$\%$\\
			& DDE & $\bm{0.173}$  & $\bm{99.47\%}$ & $\bm{53.628}$  & $\bm{97.23\%}$ & $\bm{1.047}$  & $\bm{98.88\%}$ & $\bm{0.096}$  & $\bm{97.33\%}$ & $\bm{0.034}$  & $\bm{94.22\%}$\\
			\hline
			\multirow{6}*{SEIRD} & Neural ODE \cite{12Chen}& 9.664  & 95.97$\%$ & 10858.052  & 76.20$\%$ & 39.060  & 91.66$\%$ & 0.366  & 93.27$\%$ & 1.271  & 54.99$\%$\\
			& Nelder-Mead \cite{Nelder_Mead}& 57.171  & 90.14$\%$ & 2727.172  & 82.56$\%$ & 184.118  & 85.80$\%$ & 5.298  & 29.72$\%$ & 8.083  & 40.45$\%$\\
			& Powell \cite{Powell}& 32.180  & 92.33$\%$ & 2738.686  & 82.70$\%$ & 112.195  & 87.98$\%$ & 1.546  & 40.01$\%$ & 4.001  & 45.21$\%$\\
			& BFGS \cite{BFGS}& 63.887  & 53.76$\%$ & 2657.202  & 82.03$\%$ & 70.900  & 29.45$\%$ & 8.750  & 59.88$\%$ & 54.941  & 47.91$\%$\\
			& TNC \cite{TNC}& 29.307  & 92.57$\%$ & 2611.273  & 82.26$\%$ & 32.233  & 94.00$\%$ & 6.500  & 68.70$\%$ & 3.598  & 45.98$\%$\\
			& DDE & $\bm{0.182}$  & $\bm{99.47\%}$ & $\bm{52.377}$  & $\bm{97.35\%}$ & $\bm{0.370}$  & $\bm{99.61\%}$ & $\bm{0.138}$  & $\bm{97.32\%}$ & $\bm{0.028}$  & $\bm{95.47\%}$\\
			\hline
			\multirow{6}*{SMCRD} & Neural ODE \cite{12Chen}& 30.634  & 90.77$\%$ & 13553.997  & 70.89$\%$ & 137.763  & 85.06$\%$ & 2.899  & 31.81$\%$ & 4.043  & 42.94$\%$\\
			& Nelder-Mead \cite{Nelder_Mead}& 28.687  & 1.73$\%$ & 7476.677  & 1.86$\%$ & 85.259  & 1.49$\%$ & 2.884  & 8.90$\%$ & 0.705  & 2.34$\%$\\
			& Powell \cite{Powell}& 14.957  & 96.78$\%$ & 4737.634  & 76.35$\%$ & 25.595  & 93.98$\%$ & 1.532  & 39.98$\%$ & 2.936  & 47.46$\%$\\
			& BFGS \cite{BFGS}& 28.687  & -2.78$\%$ & 7476.649  & 6.16$\%$ & 85.260  & -5.87$\%$ & 2.786  & -6.09$\%$ & 7.909  & 40.57$\%$\\
			& TNC \cite{TNC}& 24.860  & -1.62$\%$ & 7901.017  & 17.34$\%$ & 84.018  & 14.24$\%$ & 3.678  & 8.36$\%$ & 0.722  & 5.98$\%$\\
			& DDE & $\bm{0.526}$  & $\bm{98.95\%}$ & $\bm{44.209}$  & $\bm{97.15\%}$ & $\bm{2.074}$  & $\bm{98.71\%}$ & $\bm{0.027}$  & $\bm{97.42\%}$ & $\bm{0.028}$  & $\bm{95.38\%}$\\
			\hline
			\multirow{6}*{SEMCRD} & Neural ODE \cite{12Chen}& 5.475  & 92.13$\%$ & 5521.649  & N & 69.415  & 86.98$\%$ & 0.727  & 58.02$\%$ & 0.720  & 52.35$\%$\\
			& Nelder-Mead \cite{Nelder_Mead}& 16.453  & 26.34$\%$ & 5899.235  & N & 85.259  & -2.32$\%$ & 2.464  & 11.07$\%$ & 0.405  & 61.82$\%$\\
			& Powell \cite{Powell}& 16.181  & 93.47$\%$ & 4182.335  & -79.99$\%$ & 25.209  & 93.50$\%$ & 3.436  & 32.17$\%$ & 2.989  & 46.21$\%$\\
			& BFGS \cite{BFGS}& 30.886  & 92.03$\%$ & 7476.546  & N & 99.747  & 88.81$\%$ & 17.681  & 23.07$\%$ & 1.311  & 48.45$\%$\\
			& TNC \cite{TNC}& 27.640  & 92.80$\%$ & 2094.233  & N & 115.132  & 87.80$\%$ & 3.072  & 33.07$\%$ & 3.403  & 46.15$\%$\\
			& DDE & $\bm{0.459}$  & $\bm{99.17\%}$ & $\bm{41.598}$  & $\bm{87.17\%}$ & $\bm{0.386}$  & $\bm{99.48\%}$ & $\bm{0.033}$  & $\bm{96.94\%}$ & $\bm{0.014}$  & $\bm{97.59\%}$\\	
			\hline
		\end{tabular}
		\begin{tablenotes}
			\item[*] N represents the prediction results and the real data are without correlation.
		\end{tablenotes}
	\end{threeparttable}
\end{table*}

\section{Experiments and Discussions}
\subsection{Implemental Details}
In the experiment, we collect two kinds of data: countries and regions. In countries, three are used in model training and testing: the United States of America (USA), Columbia (CO), and South Africa (ZA). Moreover, two regions are considered in our model implementation: Wuhan city in China (WH) and Piedmont in Italy (PD). For data collection, we use data including the number of total accumulated infection cases, the number of disease recoveries, and the death number in \cite{Wahltinez2020}. The data include January 24$^{th}$ to April 15$^{th}$ for CN-WH and February 24$^{th}$ to June 8$^{th}$ for PD. In countries data collection, we have January 23$^{rd}$ to August 12$^{th}$ data for USA, March 6$^{th}$ to August 11$^{th}$ for CO, and March 7$^{th}$ to August 12$^{th}$ for ZA. All statistical data are collected from official notifications of various countries, WHO, National Health Commission, etc. For model testing, we used the last 20 days' data to calculate the model performance, and the last data are used in model training.

We used the experienced initial value of parameters based on \cite{17Backer} because both traditional parameter estimation methods and our DDE algorithm need initialization. According to the experiment result, the median time from onset to clinical recovery for mild cases is approximately two weeks, so we chose 0.07 as parameter $\delta_1$. Due to 3-6 weeks of recovery time for patients with the severe or critical disease, the initial $\delta_2$ is 0.03. The period from onset to the development of the severe diseases is one week according to the previous study \cite{17Backer}. 0.15 is chosen for the initial $\alpha$. The time from symptom onset to outcome ranges from 2-8 weeks among patients who have died. Thus, we consider the middle time of death time like six weeks and the initial $\varepsilon$ set to 0.03. For model with neural network (NN), we choose 0.15 of initial $\gamma$, 0.15 of initial $\alpha$, 0.07 of initial $\delta_1$, 0.03 of initial $\delta_2$ and 0.03 of $\varepsilon$. For using NN to fit $\beta$, 0.5 is used as the initial bias of NN, 0.0 is the mean value of the initial weight of NN, and 0.01 is the standard deviation (std) of the initial weight of NN. All the experiments have been implemented on Intel XeonE5-2630 v4 @ 2.20GHzz CPU and NVIDIA RTX 2080Ti GPU on ArchLinux. We implement all models in Pytorch. The DDE algorithms implementing in all epidemiological equations are trained for 5000 iterations. The learning rate (LR) is selected as 1e-3, and it will decay to 0.95$\times$lr after 400 iterations.

\begin{figure*}[htp]
	\begin{center}
		\includegraphics[width=7.0in]{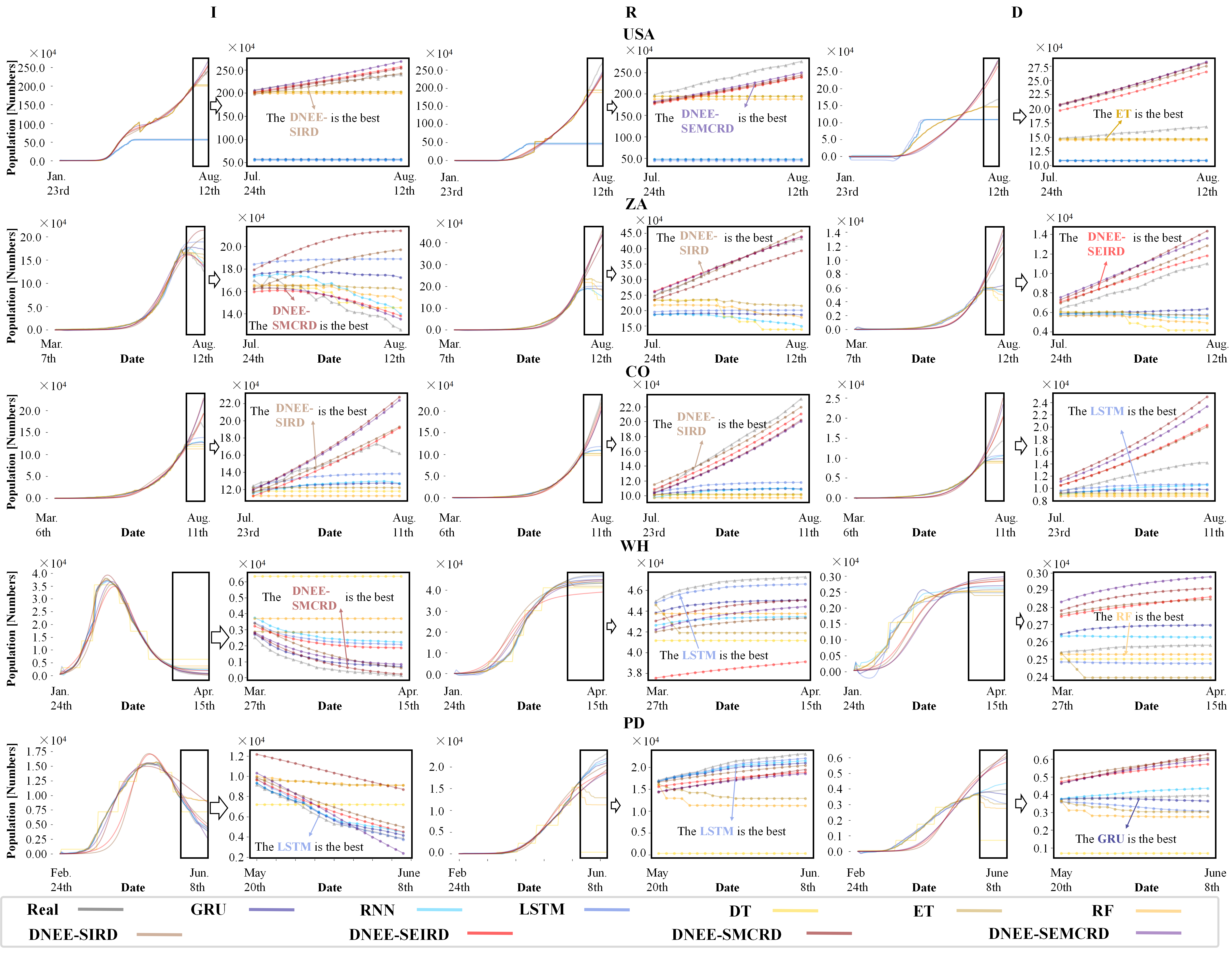}
		\caption{The figure shows all models' performance (machine-learning: RF, ET, DT, deep-learning: LSTM, RNN, GRU, our models: DDE-SIRD, DDE-SEIRD, DDE-SMCRD, DDE-SEMCRD) on all data. There are six pictures in each row in the figure, and every two pictures are a group. In a group of images, the left image represents overall trend fitting, and the right illustrates the prediction of the 20-day testing data. The group of pictures on the left represents the prediction of infection I. The group of pictures in the center represents the prediction of recovery R. The group of two pictures on the right represents death D. Furthermore, in the picture shown in the 20-day test situation, we have given the best model name corresponding to the prediction performance. From the overall predictions, our DDE model has better prediction performance in the prediction of multiple regions.}
		\label{fig:pred}
	\end{center}
\end{figure*}

To compare the fitting effects of different models and prove our DDE algorithm's advantages, we conduct several experiments. First, we implement the DDE algorithm based on two classical epidemiological equations (SIR and SEIR) and four variants designed by ourselves (SIRD, SEIRD, SMCRD, and SEMCRD). The four models of DDE algorithms are called DN-SIRD, DN-SEIRD, DN-SMCRD, and DN-SEMCRD. We study the difference between our DDE algorithm and traditional parameter estimation methods. In part B. of the experiment, we introduced the comparison results of parameter estimation (minimizing algorithms: Nelder-Mead, Powell, Truncated Newton Conjugate-Gradient (TNC)), neural ODE, and our DDE. Furthermore,  We compared the four models with the state-of-art machine learning (Decision Tree, Extremely randomized trees, Random Forest) and deep learning methods (RNN, LSTM, GRU), and comparison results are shown in part C.  

\subsection{Comparisons of traditional parameter estimation methods and DDE}
Using parameter estimation methods to assess essential parameters, including infection, mortality, and the recovery rate is crucial for SEIR and the improved SEIR model to determine these parameters. Therefore, we compared the effects of using traditional parameter estimation methods and using the DDE method we propose. As shown in Table \ref{tab:optimizedr}, we obtain the comparison performance of different models in the national data is as follows. In the USA, the MSE obtained by the best method among the traditional parameter estimation methods is 2093.261, and the best MSE of the DDE method is 41.598. The MSE obtained by the best method among the parameter estimation methods is 5.475, and the best MSE obtained by the DDE  model is 0.173. On ZA, the MSE obtained by the best method among the parameter estimation methods is 25.209, and the DDE method's MSE is 0.370.

\begin{table*}
	\caption{Model Correlation Coefficient (Pearson) of RNN, LSTM, GRU, Decision Tree (DT), Extremely randomized trees (ET), Random Forest(RF), and DDE models based on SIRD,  SEIRD, SMCRD, and SEMCRD))}
	\label{tab:corr}
	\centering
	\begin{threeparttable}
		\begin{tabular}{c|c|c|c|c|c|c|c|c|c|c|c}
			\hline
			\multirow{2}*{Areas} & \multirow{2}*{Data} & \multicolumn{10}{c}{Models}\\
			\cline{3-12}
			&& \multirow{2}*{GRU} & \multirow{2}*{LSTM} & \multirow{2}*{RNN} & \multirow{2}*{DT} &\multirow{2}*{ET} & \multirow{2}*{RF} & DDE- & DDE- & DDE- & DDE-\\
			&&     &      &     &    &    &    & SIRD & SEIRD &SMCRD &SEMCRD \\
			\hline 
			\multirow{2}*{CO} & $I$ & 92.68$\%$ & 92.96$\%$ & 94.52$\%$ &  N  &  N   &  N   &  $\bm{95.77\%}$   & 95.13$\%$ & 94.97$\%$ & 94.98$\%$\\
			& $R$ & 88.72$\%$ & 90.97$\%$ & 90.86$\%$ &  N   &  N  &  N   &  $\bm{99.86\%}$ & 99.85$\%$ & 99.78$\%$ & 99.80$\%$\\
			& $D$ & 93.85$\%$ & 95.28$\%$ & 98.51$\%$ &  N   &  N  &  N  & 98.83$\%$ & 98.77$\%$ & 98.52$\%$ &  $\bm{98.53\%}$  \\
			\hline
			\multirow{2}*{USA}& $I$ &  N   & -64.96$\%$ &  N   &  N   &  N   &  N   &  $\bm{97.81\%}$ & 97.55$\%$ & 97.54$\%$ & 97.49$\%$\\
			& $R$ &  N  & 62.31$\%$ &  N   &  N  & -87.91$\%$ &  N  &  $\bm{99.76\%}$   & 99.73$\%$ & 99.73$\%$ & 99.71$\%$\\
			& $D$ &  N  & 64.81$\%$ &  N   &  N   &  N   &  N  &  $\bm{99.71\%}$   & 99.69$\%$ & 99.69$\%$ & 99.67$\%$\\
			\hline
			\multirow{2}*{ZA}& $I$ & 72.39$\%$ & -77.85$\%$ & 95.69$\%$ & 92.24$\%$ & 79.95$\%$ & 93.10$\%$ & -95.06$\%$ & 94.83$\%$ & -91.85$\%$ &  $\bm{96.39\%}$  \\
			& $R$ & -43.67$\%$ & 88.58$\%$ & -92.86 & -92.80$\%$ & -82.24$\%$ & -90.96$\%$ & 99.35$\%$ &  $\bm{99.70\%}$   & 99.61$\%$ & 99.69$\%$\\
			& $D$ & 98.63$\%$ & 82.90$\%$ & -92.61$\%$ & -92.92$\%$ & -90.31$\%$ & -95.18$\%$ &  $\bm{99.80\%}$   & 99.66$\%$ & 99.79$\%$ & 99.73$\%$\\
			\hline
			\multirow{2}*{CN-WH}& $I$ &  $\bm{0.9985}$   & 99.79$\%$ & 99.84$\%$ &  N   & 85.52$\%$ &  N  & 98.23$\%$ & 99.78$\%$ & 98.87$\%$ & 99.46$\%$\\
			& $R$ & 99.78$\%$ &  $\bm{99.84\%}$   & 99.80$\%$ &  N  & -94.92$\%$ &  N   & 98.23$\%$ & 94.67$\%$ & 98.73$\%$ & 98.08$\%$\\
			& $D$ & 97.35$\%$ & -98.93$\%$ & -93.32$\%$ &  N   & -73.68$\%$ &  N   &  $\bm{98.89\%}$   & 96.17$\%$ & 99.17$\%$ & 98.75$\%$\\
			\hline
			\multirow{2}*{Italy-PD}& $I$ & 99.82$\%$ & 99.76$\%$ &  $\bm{99.82\%}$  & N  & 98.45$\%$ & 96.66$\%$ & 99.23$\%$ & 99.34$\%$ & 98.15$\%$ & 98.15$\%$\\
			& $R$ & 99.73$\%$ &  $\bm{99.84\%}$  & 99.78$\%$ &  N  & -94.92$\%$ & -70.83$\%$ & 99.33$\%$ & 99.41$\%$ & 98.74$\%$ & 99.44$\%$\\
			& $D$ & -85.10$\%$ & -98.71$\%$ & 99.20$\%$ &  N  & -92.77$\%$ & -75.72$\%$ &  $\bm{99.83\%}$  & 99.81 $\%$ & 99.78$\%$ & 99.77$\%$\\
			\hline
		\end{tabular}
	\end{threeparttable}
\end{table*}

Furthermore, on the regional data, the best method among the parameter estimation methods obtains an MSE of 0.405 on PD, and the best MSE of DDE models is 0.014. Also, the best method among the parameter estimation methods gets an MSE of 0.366 on CN-WH, and the best MSE of the DDE method is 0.027 on CN-WH. From this, we can find that the best result of the parameter estimation models in the USA is about 50 times that of the DDE model and the best parameter estimation method in the CO is about 32 times than DDE, about 68 times than DDE on ZA, about 29 times than DDE on PD, and about 14 times than DDE on CH-WH. Thus, our DDE method has outstanding fitting performance than traditional parameter estimation methods.

Also, as we can see in Table \ref{tab:optimizedr}, on the Pearson coefficient, the performance of the DDE method is also far better than all traditional parameter estimation methods. The best correlation in the USA's DDE method is 98.75$\%$, and that of the best method of traditional parameter estimation methods is 88.19$\%$. In CO, the best DDE method obtains 99.77$\%$ in Pearson coefficient, and in the best parameter estimation method, the Pearson coefficient is 95.97$\%$. The Pearson coefficient obtaining in ZA is 99.61$\%$ of the best DDE model, and in the traditional parameter estimation method, the Pearson value is 93.98$\%$. Moreover, in cities, the best Pearson coefficient of the best DDE in CH-WH is 98.33$\%$, and the Pearson value of the best parameter estimation method is 93.27$\%$. In Italy-PD, the best DDE gains 97.59$\%$ in Pearson value, and the best traditional parameter estimation method obtains 61.82$\%$. Therefore, we can conclude that the DDE model outperforms all traditional parameter estimation methods in correlation and precision.



\begin{figure*}[ht]
	\begin{center}
		\includegraphics[width=6.5in]{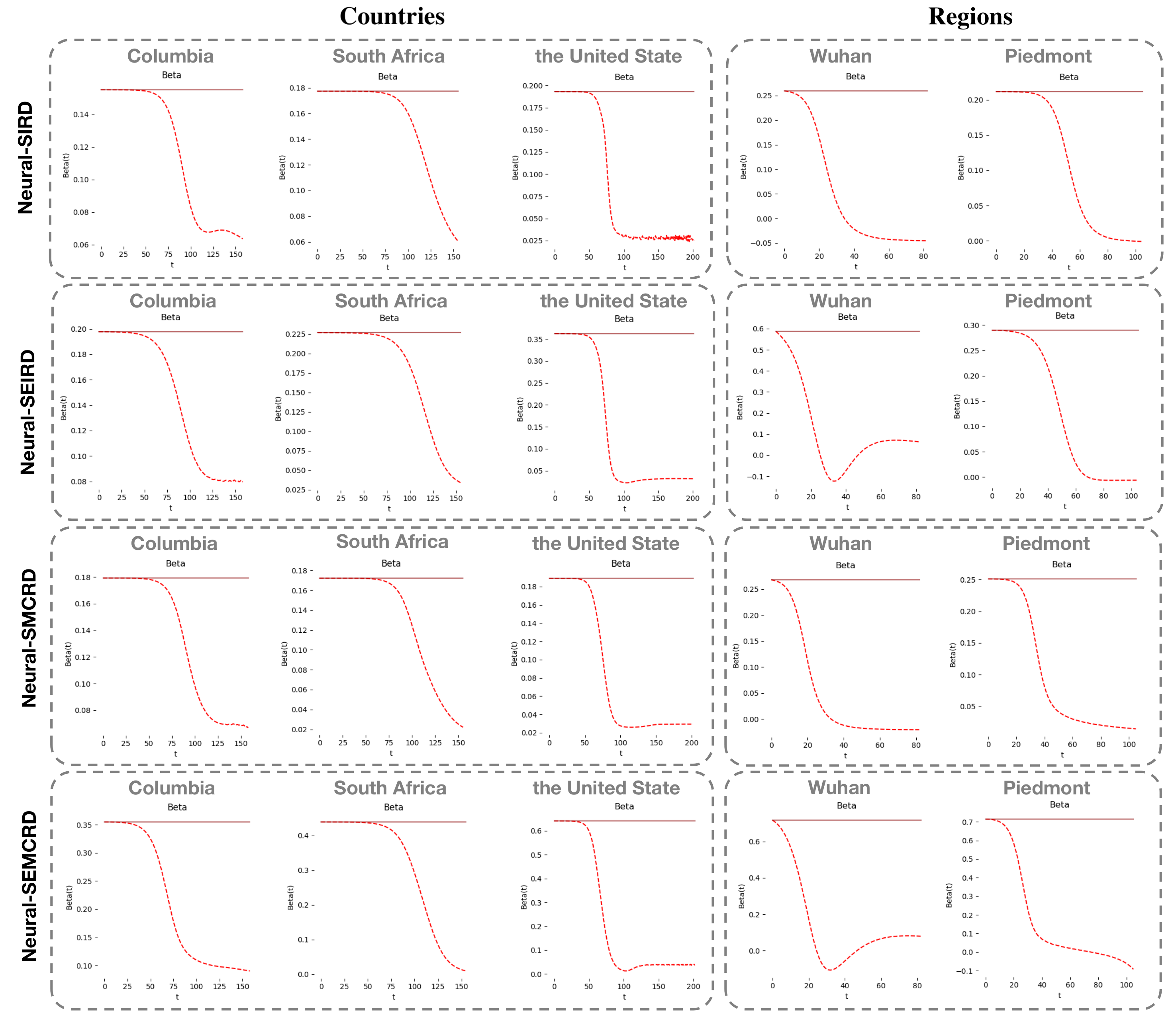}
		\caption{Schematic diagram of the infection rate beta genrated by each model and different countries and regions. Models include Neural-SIRD, Neural-SEIRD, Neural-SMCRD and Neural-SEMCRD. The dark red line in the figure represents the initial infection rate, and the red dotted curve line represents the change in the infection rate under the influence of the diversity of each country. The curves are simulated by a neural network.}
		\label{fig:pamter-NN}
	\end{center}
\end{figure*}

\subsection{Comparisons of DDE with Learning-based Methods}
To further confirm our DDE method's high fitting ability, we also compare the learning-based models with four of our DDE models (DDE-SIRD, DDE-SEIRD, DDE-SMCRD, DDE-SEMCRD). Fig. \ref{fig:pred} shows the overall forecast results on national data (CO, ZA, and the United States) and regional data (Wuhan, China, Piedmont, Italy). From the overall trend, as a representative of the deep learning method, LSTM can often achieve better results when the 20-day test data changes slowly, but the stability is not good. Sometimes, it even got the opposite trend of the real-world data, as on Piedmont's death curve. The DDE models can better capture the changing laws of the data, and some of the more complex changes are closer to the development of real data: on the infection curve in South Africa. Further, in Fig \ref{fig:pred}, the results of 20-day test data show that our DDE models obtain the best prediction performance in most situations. 

To further confirm our DDE method's high fitting ability, we also compare the learning-based models with four of our DDE models (DDE-SIRD, DDE-SEIRD, DDE-SMCRD, DDE-SEMCRD). Fig. \ref{fig:pred} shows the overall forecast results on national data (CO, ZA, and the United States) and regional data (Wuhan, China, Piedmont, Italy). From the overall trend, as a representative of the deep learning method, LSTM can often achieve better results when the 20-day test data changes slowly, but the stability is not good. Sometimes, it even got the opposite trend of the real-world data, as on Piedmont's death curve. The DDE models can better capture the changing laws of the data, and some of the more complex changes are closer to the development of real data: on the infection curve in South Africa. Further, in Fig \ref{fig:pred}, the results of 20-day test data show that our DDE models obtain the best prediction performance in most situations. 

Moreover, the 20-day test results on national data (CO, ZA, and the USA) are shown in Fig. \ref{fig:pred}. The results of USA data show that our Neural-SEIR models can better fit the data trends of $I$ and $R$, while on $D$, the growth of $D$ will be slightly overestimated. The method corresponding to deep-learning has basically no increase in $I$, $R$, and $D$, and the effect of estimating the trend is not obvious. Furthermore, on CO, the two models of Neural-SIRD and Neural-SEIRD in our model can better fit the basic trend of $I$, but some slight drops cannot be predicted. The fitting of $R$ is more accurate, but there is a certain overestimation of $D$. However, in CO, the deep-learning method still has the problem that the trend is not obvious. In the fitting of ZA, since the trend of $I$ has decreased, the difficulty of correct estimation has increased. The Neural-SIRD model and the Neural-SMCRD model have incorrect estimates of trends. The Neural-SEIRD and SECRD with NN models can accurately estimate the downward trend of $I$. Furthermore, on $R$ and $D$, all of our four models are good estimates of the trend. The deep-learning method can also better predict the downward trend on $I$, and the RNN estimate is the closest to the true value. However, in the fitting of $R$ and $D$, the deep-learning method not only fails to predict the direction of the data but obtains the opposite result from the true value.

Furthermore, the right part of Fig. \ref{fig:pred} also shows the 20-day test results on the regional data (PD and CN-WH). No matter in PD or CN-WH, all methods can basically estimate the trend of predicted data. Furthermore, some deep-learning methods have achieved better results in CN-WH and PD. Thus, the right part of Fig. \ref{fig:pred} also shows the 20-day test results on the regional data (PD and CN-WH). No matter in PD or CN-WH, all methods can basically estimate the trend of predicted data. Furthermore, some deep-learning methods have achieved better results in CN-WH and PD. It can be found that on regional data, all models have a better fitting effect; on national data, due to the complexity, the deep-learning method is not as effective as our method.

The correlation metrics including Pearson correlation coefficient (Pearson) for our DDE-SIRD, DDE-SEIRD, DDE-SMCRD, DDE-SEMCRD, and other learning-based models (deep-learning method: RNN, LSTM, GRU and traditional machine learning method: Random Forest, Extra Tree, Decision Tree) of 20-day test data are shown in Table \ref{tab:corr}. According to the Pearson metric for real-world data in Table \ref{tab:corr}, deep learning methods such as LSTM and RNN models have achieved good results in regional data fitting. In the data fitting of CN-WH, the best Pearson on $I$ and $R$ are obtained by GRU and LSTM respectively. The situation of PD data is similar. The best Pearson for $I$ and $R$ are obtained on RNN and LSTM respectively. However, there are still unstable problems in the fitting of these deep learning algorithms, and the negative correlation between the predicted data and the real data often occurs. 

In addition, machine learning methods including random forest, extra tree, and decision tree are not very effective in fitting regional data. There is a situation where the prediction has no correlation or the correlation is negative with real data. However, our DDE models (DDE-SIRD, DDE-SEIRD, DDE-SMCRD, and DDE-SEMCRD) have achieved relatively stable and excellent performance in the overall prediction of regional $I$, $R$, and $D$ data. Based on the Wuhan data, the four deformed Neural-SEIR models have achieved $I$, $R$, and $D$ average correlation coefficients of 98.45$\%$, 97.87$\%$, 98.92$\%$, and 98.76$\%$ respectively. Furthermore, on the PD data, they also obtained the average Pearson correlation coefficients of 99.46$\%$, 99.52$\%$, 98.56$\%$, and 99.12$\%$ respectively.

\subsection{The estimated paramaters of DDE models}
In the USA, the average initial infection rate $\beta *$ is 0.469, and we obtain the average initial infection rate $\beta *$ of 0.222 in the CO. The average initial infection rate $\beta *$ is 0.254 in the ZA. In regions, the average $\beta *$ in Wuhan is 0.4578, and Piedmont gains the average initial infection rate of 0.367. The infection rates our DDE method stimulates are suitable for real-world situations. In Fig. \ref{fig:pamter-NN}, the DDE also shows that it can simulate various infection rate changes using neural networks to fit the effect function. The DDE models reflect the basic trend of the gradual decline in the infection rate and more accurately estimate the initial infection rate.

\section{Conclusions}
Our work shows that DDE is a valuable method for epidemic disease data fitting and analysis. The experimental results illustrate that traditional parameter estimation methods using an epidemiological equation to describe the development of epidemic disease face the challenge of fitting real-world data. Different interventions during the development period cannot achieve good results using unchanged and straightforward epidemiological parameters. Although the learning-based models, including LSTM and random forest, have excellent data fitting performance, the interpretability is low. Therefore, our DDE method combines simple parameters with dynamic parameters simulated by the neural networks and uses the Neural ODE method to better fit data. From the perspective of epidemiological equations, the SEMCRD model we proposed better refines the different situations of the group $I$ and group $R$, which is better for data analysis. Our model has achieved better results from the model's accuracy and correlation with the real-world data. Furthermore, our method provides new ideas and tools for future data analysis of infectious diseases.

\bibliographystyle{IEEEtran}
\bibliography{ms}

\end{document}